\documentclass[preprint,12pt]{elsarticle}
\usepackage{amsmath}

\usepackage{amssymb}

\journal{Medical Image Analysis}

\begin{document}

\begin{frontmatter}



\title{Automated segmentation of the pulmonary arteries in low-dose CT by vessel tracking}


\author{Jeremiah~Wala, Sergei Fotin, Jaesung Lee, Artit Jirapatnakul, Alberto Biancardi and~Anthony~Reeves}

\address{Department of Electrical and Computer Engineering, Cornell University, Ithaca, NY 14853}

\begin{abstract}
We present a fully automated method for top-down segmentation of the pulmonary arterial tree in low-dose thoracic CT images. The main basal pulmonary arteries are identified near the lung hilum by searching for candidate vessels adjacent to known airways, identified by our previously reported airway segmentation method. Model cylinders are iteratively fit to the vessels to track them into the lungs. Vessel bifurcations are detected by measuring the rate of change of vessel radii, and child vessels are segmented by initiating new trackers at bifurcation points. Validation is accomplished using our novel sparse surface (SS) evaluation metric. The SS metric was designed to quantify the magnitude of the segmentation error per vessel while significantly decreasing the manual marking burden for the human user. A total of 210 arteries and 205 veins were manually marked across seven test cases. 134/210 arteries were correctly segmented, with a specificity for arteries of 90\%, and average segmentation error of 0.15 mm. This fully-automated segmentation is a promising method for improving lung nodule detection in low-dose CT screening scans, by separating vessels from surrounding iso-intensity objects.
\end{abstract}

\begin{keyword}

pulmonary artery segmentation \sep vessel tracking \sep computed tomography (CT) \sep validation metric



\end{keyword}

\end{frontmatter}

\section{Introduction}

The pulmonary arteries comprise the vessels that supply deoxygenated blood to the lungs for gas exchange. Computer segmentation of the pulmonary vessels in computed tomography (CT) images has been proposed as a method for improving automated lung nodule detection in low-dose lung cancer screening scans~\cite{Croisille:1995}. Additionally, computer segmentation of the pulmonary vessels has been explored as a method to reduce false positives in pulmonary embolism detection by radiologists~\cite{Masutani:2002}, and improve pulmonary artery size measurements for pulmonary arterial hypertension patients~\cite{Linguraru:2010}. 

The pulmonary vasculature is a highly complex branching structure that presents significant challenges towards robust segmentation. Normal anatomical variability, as well as anatomic abnormality from disease, limits the amount of {\it a priori} information that can be used to develop a robust segmentation algorithm. In low-dose CT lung cancer screening scans, there is the added challenge of a high signal-to-noise ratio compared with standard diagnostic CT scans. Additionally, lung cancer screening scans do not use intravenous radiocontrast, providing low levels of contrast between the different types of dense (non-fat) soft tissue, such as arteries, veins, and lung nodules.

We present a model-based automated method for segmenting the pulmonary arteries in low-dose CT scans and separating them from surrounding isointensity objects. Seed-points for initiating a cylindrical vessel tracker are automatically located in the basal pulmonary arteries by searching near airways identified using our automated airway segmentation algorithm~\cite{LeeJ:2008}. The vessel tracker iteratively follows the arteries into the lung parenchyma, and vessel branching is detected by means of a bifurcation detector based on the rate of change of the vessel radii. 

Attempts to segment the pulmonary vasculature in CT have included region-growing~\cite{Bulow:2005, Wood:1995, Masutani:2001}, vessel enhancement with a Hessian matrix~\cite{Zhou:2007, Shikata:2009, ZhouJ:2007}, and fuzzy connectivity~\cite{Kaftan_2:2008}. Similar techniques have also been used to segment arteries and veins separately. Lei et al.~\cite{Lei:2001} used multiseeded fuzzy connectivity to separate a conflated A/V tree in contrast-enhanced magnetic resonance angiography of the abdomen. Bemmel \emph{et al.} implemented a level-set algorithm to separate arteries and veins in magnetic-resonance images~\cite{Bemmel:2003}. Recently, Saha et al.~\cite{Saha:2010} presented an approach for separating fused isointensity objects, including pulmonary arteries and veins, which utilized multiscale morphologic filtering, and fuzzy connectivity maps generated from manual seed points. 

Although previously-proposed pulmonary artery segmentation methods require the use of manually selected seed-points, in this work the segmentation is completely automated. This serves to eliminate inter-user differences, and would allow the algorithm to be included in fully-automated lung nodule detection systems. Additionally, the method presented here utilizes a cylindrical vessel tracker, which reconstructs vessels by progressively fitting model cylinders. Using a model-fitting approach on low-dose scans is advantageous because it can identify differences in geometry between two isointensity structures, even when local amounts of noise are high.

We also present a novel evaluation metric, the sparse surface (SS) metric, for use with segmentations of vascular trees. The SS metric was designed to quantify both the overall extent (rate of correctly segmented vessels) and accuracy per identified vessel in the overall segmentation. The validation scheme for the SS metric requires a manageably sparse number of manual markings, increasing the number of cases that experimental algorithms can be validated on in a limited amount of time.  

\section{Materials and methods}
\subsection{Preliminary steps}
Three preliminary steps were performed to reduce noise and decrease the signal from unwanted structures. A $3\times3$ median filter was first applied to reduce the non-Gaussian noise inherent in low-dose scans. The scans were then thresholded at -400 HU to separate the soft tissue (0 HU) from the lung parenchyma (-900 HU) within the lungs. Structures such as bone (200 HU) and fat (50 HU) were not eliminated after thresholding. However, because these structures do not appear in the lungs, they are highly unlikely to affect the results of the segmentation. 

\begin{figure}[h]
\begin{center}$
\begin{array}{cc}
\includegraphics[width=4.3in]{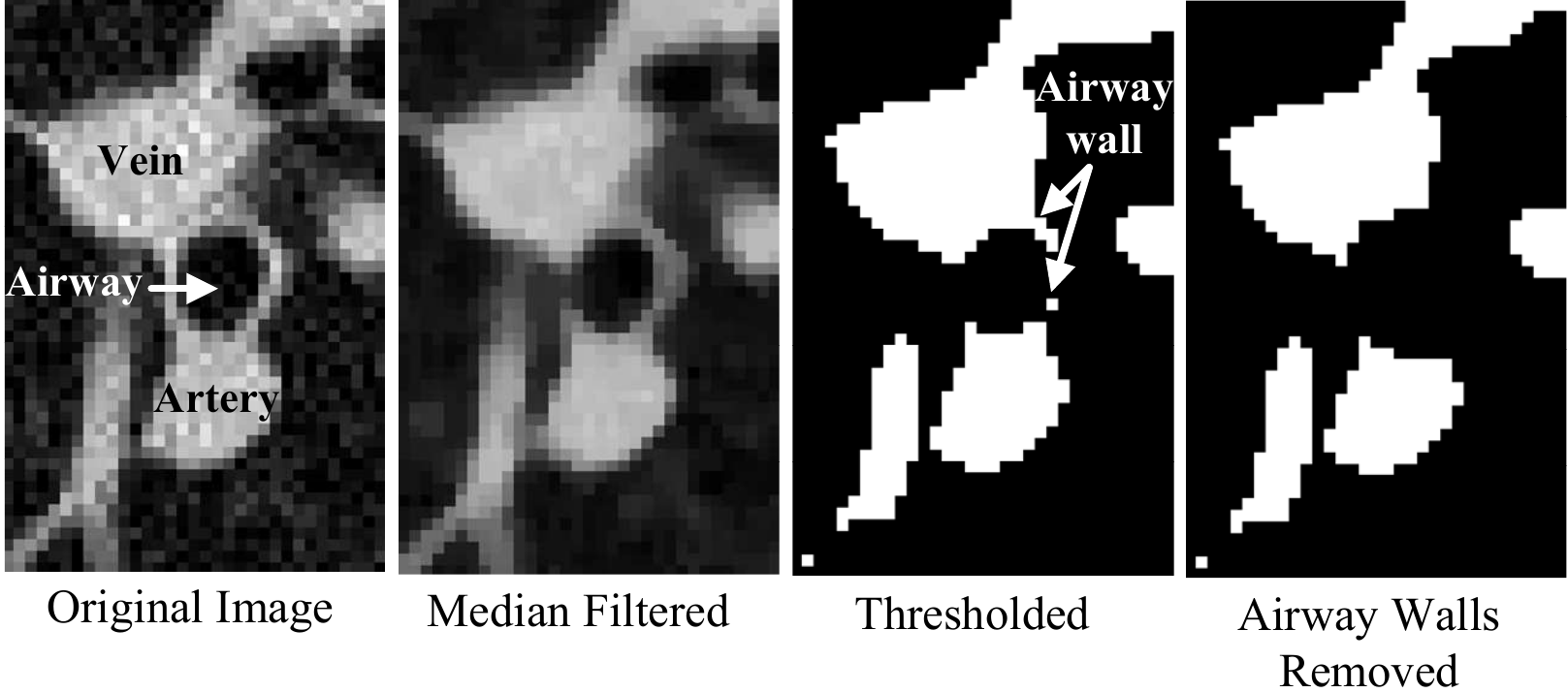} 
\end{array}$
\end{center}
\vspace{-23pt}
\caption{Removal of the airway walls by dilation of the airway segmentation and subtraction from the thresholded CT image.}
\vspace{-10pt}
\label{fig:airelim}
\end{figure}

We used the results of our previously reported airway segmentation~\cite{Lee:2009} to remove the airway walls, a potential source of error in vessel segmentation. The airway was enlarged by dilation with a 2 mm isotropic kernel. The dilated airway was then subtracted from the CT image to remove the airway walls. Figure~\ref{fig:airelim} illustrates the application of this operation to a 3rd generation airway/artery segment.

\subsection{Automated seed point detection}

\begin{figure}[h]
\begin{center}$
\begin{array}{cc}
\includegraphics[width=5.3in]{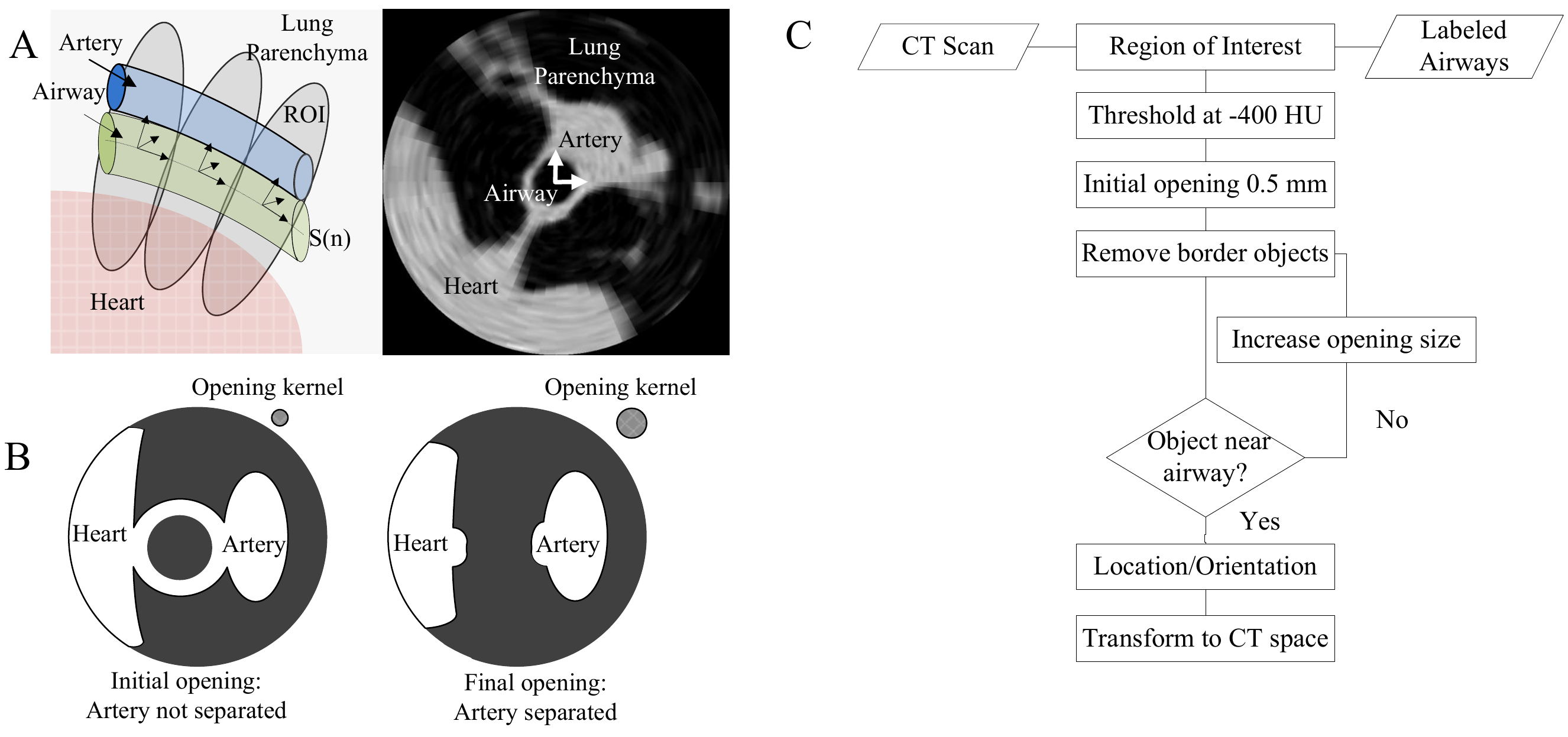} 
\end{array}$
\end{center}
\vspace{-15pt}
\caption{a) Diagram of the airway ROI construction process and a single frame from the output. b) Progressive morphological opening operation to isolate the pulmonary artery. c) Complete process diagram for the identification of seed points within the pulmonary artery.}
\vspace{-10pt}
\label{fig:pseed}
\end{figure}

The airways and pulmonary arteries run parallel and adjacent to each other in units known as bronchopulmonary segments. This symmetry provides the key for our automated pulmonary artery seed point detector. The airway lumen is automatically segmented and labelled using a region-growing method with an automatically generated seed point in the trachea~\cite{Lee:2009}. Using the segmented airway, a curved cylindrical region of interest is constructed around individual airway branches to limit the search space for the pulmonary arteries. A cubic spline $S(n)$ is fit to the airway, and the scan is resampled to construct cutting planes orthogonal to the spline, spaced at constant intervals. Figure~\ref{fig:pseed}a diagrams the processes of generating the 3D ROI, and shows a single frame for illustration.
 
The algorithm for locating the pulmonary arteries within the reconstructed ROI is outlined in Figure~\ref{fig:pseed}c. The ROI is thresholded at -400 HU, and the soft tissue structures are separated using a 3D morphological opening operation to separate the heart from the pulmonary arteries (Figure~\ref{fig:pseed}b). The resulting objects are labeled with a 3D connected component algorithm.

Each unique object is then checked to see if it is a possible candidate for the pulmonary artery. Objects which extend past 4 cm from the center of the airway contain areas of the heart and are eliminated. Of the remaining objects after the elimination, the component that is closest to the airway center is identified as the pulmonary artery, as predicted by the bronchopulmonary model. If no components are found, the morphological operation opening was too small to isolate the artery. In this case, the size of the kernel is increased by one voxel and the process is repeated until the pulmonary artery is identified.

The location of the isolated pulmonary artery is then defined as the centroid of the artery, in the ROI frame most proximal to the heart. This ensures that vessel tracking begins as close to the start of the vessel as possible, increasing the extent of the segmentation. The orientation of the artery was found by computing the centroid of the artery in the most distal frame it appears in the ROI, and taking the vector from the proximal centroid to the distal centroid as the initial artery orientation. 

\subsection{Vessel tracking}

A single vessel can be modeled as a series of discrete cylinders based on the observation that they are approximately cylindrical in shape. Vessel tracking is the task of recreating the cylindrical vessel by progressively fitting model cylinders in the CT space.

The similarity metric between a model cylinder and the CT scan is constructed from the cylinder $C$ and set $V$ of voxels in the thresholded image. The total similarity score $S$ is the sum of scores $s_i$ of all voxels $v_i$ in $C$:
\begin{equation}
 S = \displaystyle\sum_{i}s_i \quad \text{where} \quad s_i = \begin{cases}1 &\text{if}\;\; v_i > 0 \\ -5 &\text{if}\;\;v_i = 0\end{cases}\quad\forall v_i \in C
\end{equation} 
The strict penalty of $-5$ was selected {\it ad hoc} to prevent the cylinder tracker from jumping between adjacent soft tissue structures -- model cylinders that span areas of lung parenchyma are highly penalized. The similarity metric is not normalized to the size of the cylinder, which rewards cylinders that capture more of the desired vessel.

\begin{figure}[h]
\begin{center}$
\begin{array}{cc}
\includegraphics[width=5.3in]{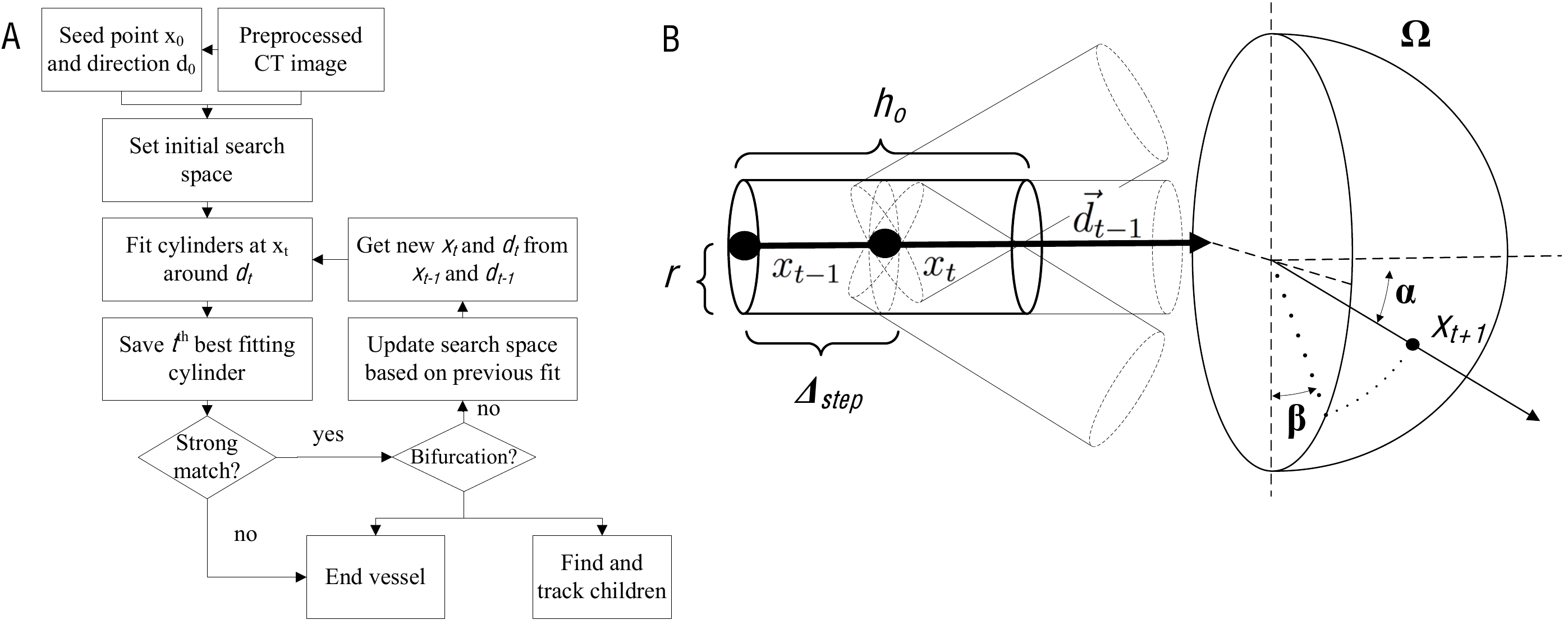} 
\end{array}$
\end{center}
\vspace{-15pt}
\caption{a) Process diagram for iterative vessel tracking. Starting from an initial seed point $x_0$ and direction $\hat{d}_0$, cylinders are iteratively fit to the image, recreating the vessel. New vessels are initiated at points of bifurcation. b) Diagram showing the progressive fitting of model cylinders. Several model cylinders of varying radii and orientation are checked for their similarity to the vessel.}
\label{fig:cylfit}
\end{figure}

The iterative vessel tracking process is outlined in Figure~\ref{fig:cylfit}a. Beginning at seed point $x_0$ and direction $\hat{d}_0$, model cylinders of varying orientations and radii are fit to the vessel. The initial radius $r_0$ is optimized on the interval [2 mm, 12 mm], to cover both abnormally large and small vessels. The orientation search space is limited to the solid angle $\Omega = 2\pi$, which allows for sharp turns in vessels. To limit the size of the model parameter space, the cylinder height is set to a fixed value of $h_0$. The segmented vessel is then the set of soft-tissue pixels contained within the best fitting model cylinder.

After the initial model cylinder is fit, the process repeats for a new initial position $x_1$, shown in Figure~\ref{fig:cylfit}b. For the $t^{\text{th}}$ iteration of cylinder fitting, the initial point $x_t$ is defined as:
\begin{equation}
x_t = (\Delta_{\text{step}}) \hat{d}_{t-1} + x_{t-1}
\end{equation}
where $\hat{d}_{t-1}$ is the normalized direction of the previous model cylinder, and $\Delta_\text{step}$ is the distance the next cylinder is progressed from the previous one. Because the radius is expected to vary slowly between model cylinders, the radius search space is limited on:
\begin{equation}
r_{t} \in [0.7 r_{t-1}, 1.3 r_{t-1}]
\end{equation}

Iterative vessel tracking terminates when the tracker either fails to find a strong match or a bifurcation is detected. A strong match is defined as having at least 50\% soft-tissue voxels within the model cylinder. This criterion fails to be met when either the vessel becomes too small to give a strong signal, as in the case of higher generational subsegmental arteries, or the tracker has fallen off the vessel.

\subsection{Bifurcation detection}


To initiate tracking of child vessels, a method was developed to search for child branches and initiate new vessel trackers. Automated bifurcation detection is performed by comparing the actual vessel geometry to a bifurcation model based on the following observations:
\begin{enumerate}
\item The radii of the child vessels is smaller than the radii of their parent vessel.
\item The angle between child vessels is no less than $30^\text{o}$ and no more than $90^\text{o}$. 
\item The child vessels curve only slowly away from the parent vessel.
\end{enumerate}

\begin{figure}[h]
\begin{center}$
\begin{array}{cc}
\includegraphics[width=5.3in]{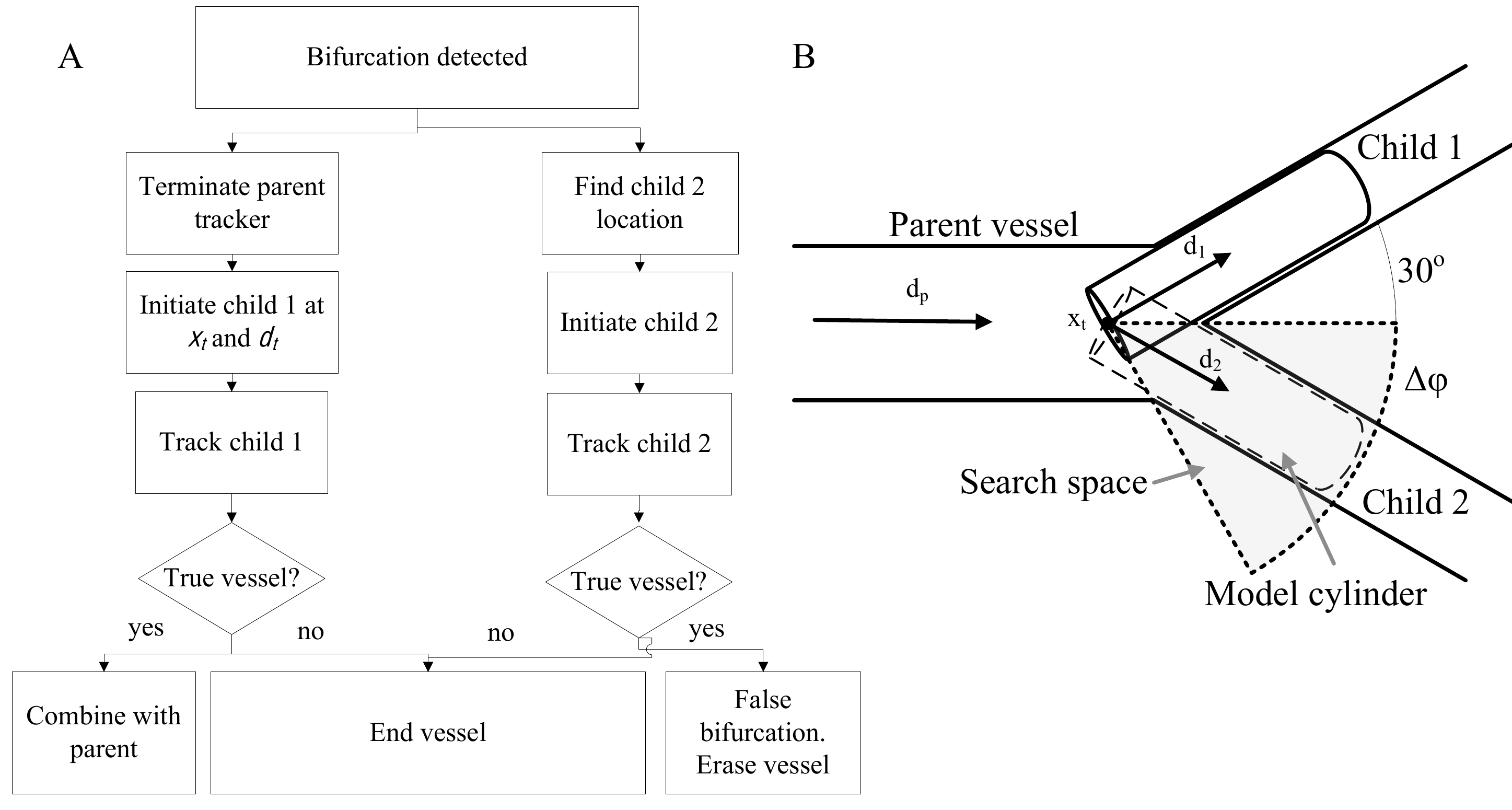} 
\end{array}$
\end{center}
\vspace{-15pt}
\caption{a) Process diagram for locating child vessels at bifurcation point. b) Detection of the second child vessel by searching for the optimally fitting model cylinder. The search space is limited by the orientations of the parent vessel and child 1.}
\label{fig:bifrad2}
\end{figure}

As the parent tracker encounters a bifurcation, the next model cylinder will fit onto the child vessel with a smaller radius as shown in Figure~\ref{fig:bifrad2}b. 
Using this model, a bifurcation candidate is generated when the ratio of the current radius to the radius at $\frac{1}{2}h_0$ upstream, $r_{0} / r_{\frac{h_0}{2}}$, falls below the radius change threshold $\delta_{\text{radius}}$.


The process for finding child vessels at bifurcation candidate points is outlined in Figure~\ref{fig:bifrad2}a. At candidate point $x_t$, a new child vessel (child 1) is initiated at $x_t$ in the direction $\vec{d}_t$. The next step is to search for the second child vessel. Observations 2 and 3 of the bifurcation model indicate where to look for bifurcation candidates. Let the orientation of the parent vessel at the bifurcation be $\hat{d}_p$, and the orientation of children 1 and 2 at the bifurcation be $\hat{d}_1$ and $\hat{d}_2$ respectively. Child vessel 2 will satisfy the bifurcation model if:
\begin{equation}
\cos \frac{\pi}{6} < \left(\hat{d}_1 \cdot \hat{d}_2\right) < \cos \frac{\pi}{2}  \quad \text{and} \quad \left(\hat{d}_2 \cdot \hat{d}_p\right) < \cos  \frac{\pi}{2} 
\end{equation}
The search space that satisfies these conditions is illustrated in Figure~\ref{fig:bifrad2}.

\begin{figure}[h]
\begin{center}$
\begin{array}{cc}
\includegraphics[width=4.3in]{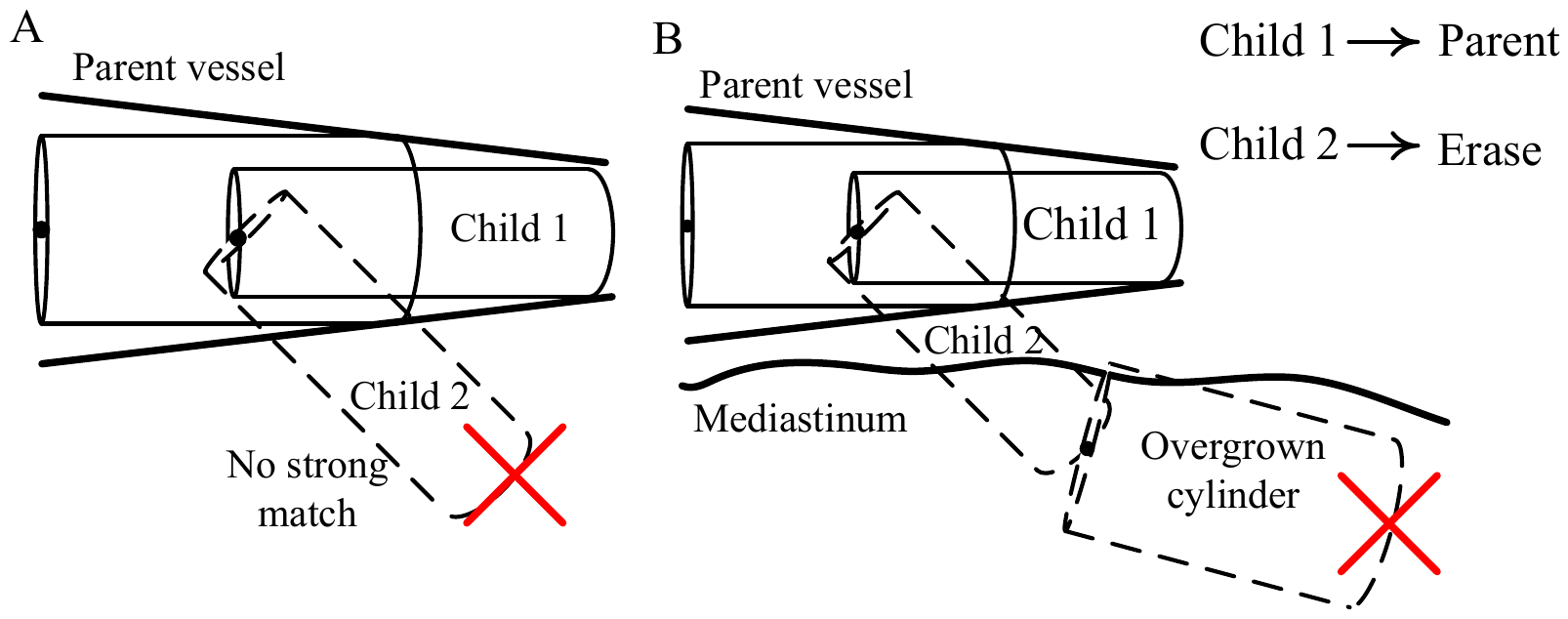} 
\end{array}$
\end{center}
\vspace{-22pt}
\caption{False positive bifurcation detection and subsequent error correcting. a) The child vessel does not find any strong matches, indicating that the bifurcation detector returned a false positive. b) The vessel has leaked into the mediastium, but is removed as the optimized radius grows too quickly.}
\label{fig:bifrad3}
\end{figure}

In the case of a false positive in the bifurcation detection, child 1 may terminate after only a few iterations. This situation is presented in Figure~\ref{fig:bifrad3}a. To correct the error generated by the false bifurcation detection, child 1 is erased and child 2 continues to track as a continuation of the parent vessel. Additionally, a leak detector was implemented to remove leaks into the nearby mediastinum (heart and great vessels), illustrated in Figure~\ref{fig:bifrad3}b. A leak is detected when a model cylinder is fit with a radius over 150\% of the radius of the model cylinder beginning at one full cylinder length upstream. 

\subsection{Sparse Surface Metric}

We developed a novel surface-based method called the sparse surface (SS) metric for validating the results of our pulmonary artery segmentation method. The goals in developing a new validation criteria to be applied to pulmonary artery segmentation were three-fold:
\begin{enumerate}
\item To quantify the number of vessels segmented (i.e. extent of segmentation).
\item To use the same markings to evaluate the error per vessel (i.e. the accuracy of segmentation).
\item To minimize the number of manual ground truth markings (points defining true object geometry) required from the human user.
\end{enumerate} 


The SS method accomplishes the three listed goals by defining a mapping from ground truth points marked at the surface of vessels to the surface of the segmentation. By mapping from the ground truth to the segmentation, ground truth can be marked only sparsely -- there is no penalty incurred by segmentation surface points that are not close to ground truth points. The full geometry of the vascular tree can be captured by marking a limited number of surface points for each vessel, allowing for the calculation of the extent of the segmentation. The following section formally defines the SS metric.

\subsubsection{Formal definition}
Let $V$ be the set of voxels of the segmented volume. The boundary of $V$ is the set $\partial V$ of voxels that are 6-connected to a background voxel. The set $S$ of surface points of a segmented volume $V$ is then defined as:
\begin{equation}
S = \{x|x\in \partial V \} = \{s_1, s_2, ..., s_k\}
\end{equation}
Let $G$ be the set of ground-truth points, where the members $g_i$ were individually identified by manual markings. The set $G$ contains $n$ subsets $O_n$ representing the ground-truth markings for $n$ objects. The SS metric is constructed from a mapping $f$ of $O_n$ to $M_n$ such that for every element $o_i$ of $O_n$:
\begin{equation}
f(o_i) = m_i =a_i\min_{j = 1, ..., k}\{ \| (o_i,s_j) \|_E \} 
\end{equation}
where the parameter $a_i$ describes whether the point is interior or exterior to the segmentation:
\begin{equation}
a_i = \begin{cases}-1\quad\text{   if } m_i \in V\quad\text{(interior)}\\\;\;\;1 \quad\text{ otherwise}  \quad \text{(exterior)}\end{cases}
\end{equation}
and $\|(a,b)\|_E$ is the positive semi-definite 3D Euclidean distance between two points:
\begin{equation}
\|(a,b)\|_E = \sqrt{\left(a_x-b_x\right)^2+\left(a_y-b_y\right)^2+\left(a_z-b_z\right)^2}.
\end{equation}

The set $M_n$ now holds the values of the Euclidean distance from each ground truth point for object $n$ to the nearest surface point of the entire segmentation, with positive values corresponding to ground truth points that lie outside the segmentation and negative values for interior points. 

The distribution of distance elements in each of the sets $M_n$ is expected to take on a normal unimodal distribution. The ideal algorithm will return a narrow distribution centered around zero. For an over-segmentation, the distribution will be shifted towards negative distances, and for an under-segmentation, will be shifted towards positive distances. 

Quantitative information about the segmentation can be obtained by computing statistics on $M_n$. The average signed distance (ASD) $\overline{M_n}$ of the set $M_n$ of order $N$ can be computed as:
\begin{equation}
\overline{M_n} =\displaystyle\sum_i^N \frac{m_i}{N}
\end{equation}
The average signed distance is a measure of the overall bias of the segmentation -- a positive ASD indicates under-segmentation, and a negative ASD indicates over-segmentation. To evaluate the magnitude of the average error per truth point, the root mean square distance (RMSD) can be calculated for each object distance error set $M_n$.
\begin{equation}
M_\text{nRMS} = \sqrt{\displaystyle\sum_i^N\frac{m_i^2}{N}}.
\end{equation}
Additionally, the maximum amount of over- and under-segmentation can be found by calculating the maximum negative distance and maximum positive distance, respectively. Table~\ref{table:stat} gives a summary of the useful statistics in the SS metric.
\begin{table}[t]
\begin{center}
\caption{Statistical measurements used in the SS evaluation metric}
\label{table:stat}
\begin{tabular}{l  l}
Statistic & Measure of \\ 
\hline\hline 
Average signed distance (ASD) & Bias of segmentation (Accuracy) \\ 
Root mean square distance (RMSD) & Magnitude of error (Precision) \\
Max negative distance & Point of greatest over-segmentation \\
Max positive distance & Point of greatest under-segmentation \\
\hline
\end{tabular}
\end{center}
\end{table}

Given a set of surface markings for each object, the distance error statistics obtained using the SS metric can be used to achieve our goals of evaluating the number of objects segmented, and the error per object. An object may be classified as having been segmented if the RMS distance error is small, signifying that the segmentation did not stray far from the object. The bias, magnitude, and extreme values of the error can then be calculated for each segmented object.

\subsubsection{Application of the SS metric to pulmonary artery segmentation}

Ground truth points were marked on axial slices of CT scans windowed to soft tissue. To automatically detect segmentation errors, veins were marked in addition to arteries. This allowed for quantification of the main source of in our segmentation method. An approximately homogenous mat of points was created by marking vessels at roughly 4 mm intervals around the circumference of arteries for roughly every third frame, corresponding to about a 4 mm interval in the axial direction. Each vessels was marked separately and given a unique label. The ASD, RMS, and maximum and minimum of the distances from the truth points to the nearest surface point were computed separately for each vessel.

Each marked vessel was classified as either having been segmented or missed by measuring the RMS error per vessel. A vessel was classified as having been segmented if the RMS the distance was below 2.0 mm, which corresponds to an average error of roughly 3-4 voxels. The threshold of 2.0 mm is an {\it ad hoc} value that represents the upper error bound for vessels visually confirmed to be segmented. Using this method, each vessel was classified into one of four categories, given Table~\ref{table:cat}, for evaluating the performance of the algorithm.

\begin{table}[h]
\begin{center}
\caption{Vessel classification for evaluating algorithm performance}
\label{table:cat}
\begin{tabular}{l l l}
Category & Type of vessel & Criterion \\
\hline \hline
True Positive (TP) & Artery & RMS error less than 2.0 mm \\
False Negative (FN) & Artery & RMS error greater than 2.0 mm \\
True Negative (TN) & Vein & RMS error greater than 2.0 mm \\
False Positive (FP) & Vein & RMS error less than 2.0 mm \\
\hline
\end{tabular} 
\end{center}
\end{table}

\subsection{Dataset}
The dataset used for this experiment consisted of 10 low-dose thoracic CT scans with correctly segmented and labeled airways. The scans were generated with an 8 slice helical CT scanner operating under a low-dose protocol with beam current of 40 mA and 120 kVP operating voltage. The slice thickness was 1.25 mm and the in-plane resolution ranged from 0.55 mm to 0.82 mm. For each case, the left and right basal artery and all child branches with radii greater than 1 mm were documented with sparse markings. In addition, all veins with radii greater than 1 mm surrounding the basal arterial tree were documented. A total of 210 arteries and 205 veins were marked across the ten cases.

\section{Results}

\subsection{Training and optimization}
Three documented cases were selected for automated algorithm training and parameter optimization. Table~\ref{tab:training} summarizes the parameters that were optimized. 

\begin{table}[h]
\begin{center}
\caption{Parameter sets used in algorithm training}
\label{tab:training}
\begin{tabular}{c c c c}
Parameter & Range & Step Size & Optimized value  \\ 
\hline\hline 
$h_o$ & 5 - 30 mm & 5 mm & 15 mm \\
$\Delta_{\text{step}}$& 0.10 - 0.30 & 0.05 & 0.20 \\
$\delta_{\text{radius}}$ & 0.75 - 1.00 & 0.05 & 0.90 \\ 
\hline
\end{tabular}
\end{center}
\end{table}
The sparse surface metric was used to automatically calculate a training score $S$ for maximization:
\begin{equation}
S = TP + TN - FP - FN
\end{equation}

\begin{figure}[h] \centering \includegraphics[width=5.3in]{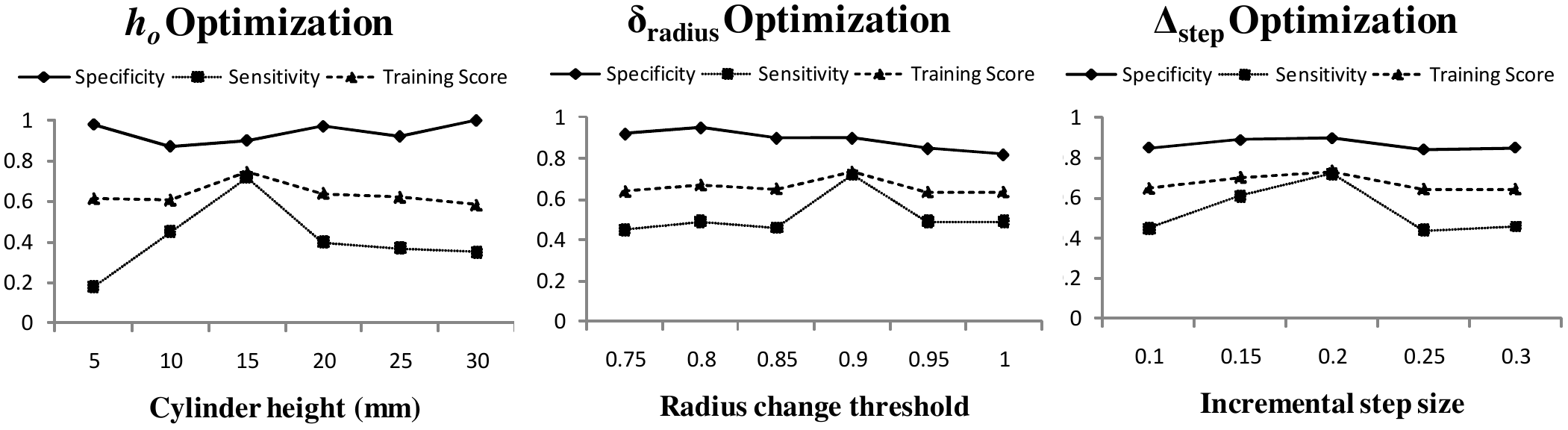} \caption{Results of the vessel tracking method for different sets of parameters. The optimal parameters were found by maximizing the training score. The optimal value for the radius change threshold $\delta_{\text{radius}}$ reflects the approximate physiological value.} \label{fig:results} \end{figure}

The parameter that had the largest effect on the outcome was the cylinder height $h_0$. Visual inspection of the training results indicate that short cylinders are highly sensitive to noise and are unable to extract the vessel structure. Conversely, cylinders that were too large were not sensitive enough to detect most bifurcations. Similarly, incremental step sizes above the optimal value of 0.20$h_0$ were too coarse to detect many bifurcations, decreasing the sensitivity.

The radius change threshold parameter $\delta_{\text{radius}}$ was optimized to a value of 0.90, which gave a marked improvement in algorithm performance over a much weaker value of 1.0. Limiting the search for child vessels to geometries where the radius abruptly changed improved overall vessel identification performance.

\subsection{Vessel tracking}
Of the 20 arterial trees to be segmented across the ten cases (10 left and 10 right arterial trees), 19 were successfully seeded as the arteries entered the lungs. The vessel trackers stemming from the automated seed point segmented a total of 134 of the 210 manually identified arteries to within an RMS distance error of less than 2.0 mm, yielding an overall sensitivity of 64\%. 20 of 205 veins were erroneously segmented as arteries, yielding an overall specificity for artery segmentation of 90\%. Figure~\ref{fig:results} shows the segmentation results, both overlaid on an axial CT slice and as a 3D surface representation.

In most cases, the tracker was able to separate the arteries from the veins by searching for cylindrical objects in a limited size and orientation search space defined by the bifurcation model. While this approach was able to distinguish nearly identical vessels based only on their relationship to other arteries, it is unable to locate vessels that fall outside of the search space. Thus, arteries were missed if they originated at apparent trifurcations, or had unusually small radii relative to their sister branches. Relaxing the bifurcation model results in few missed segments, but more false positives (i.e. segmented veins). 

\begin{figure}[h]
\begin{center}$
\begin{array}{cc}
\includegraphics[width=5.3in]{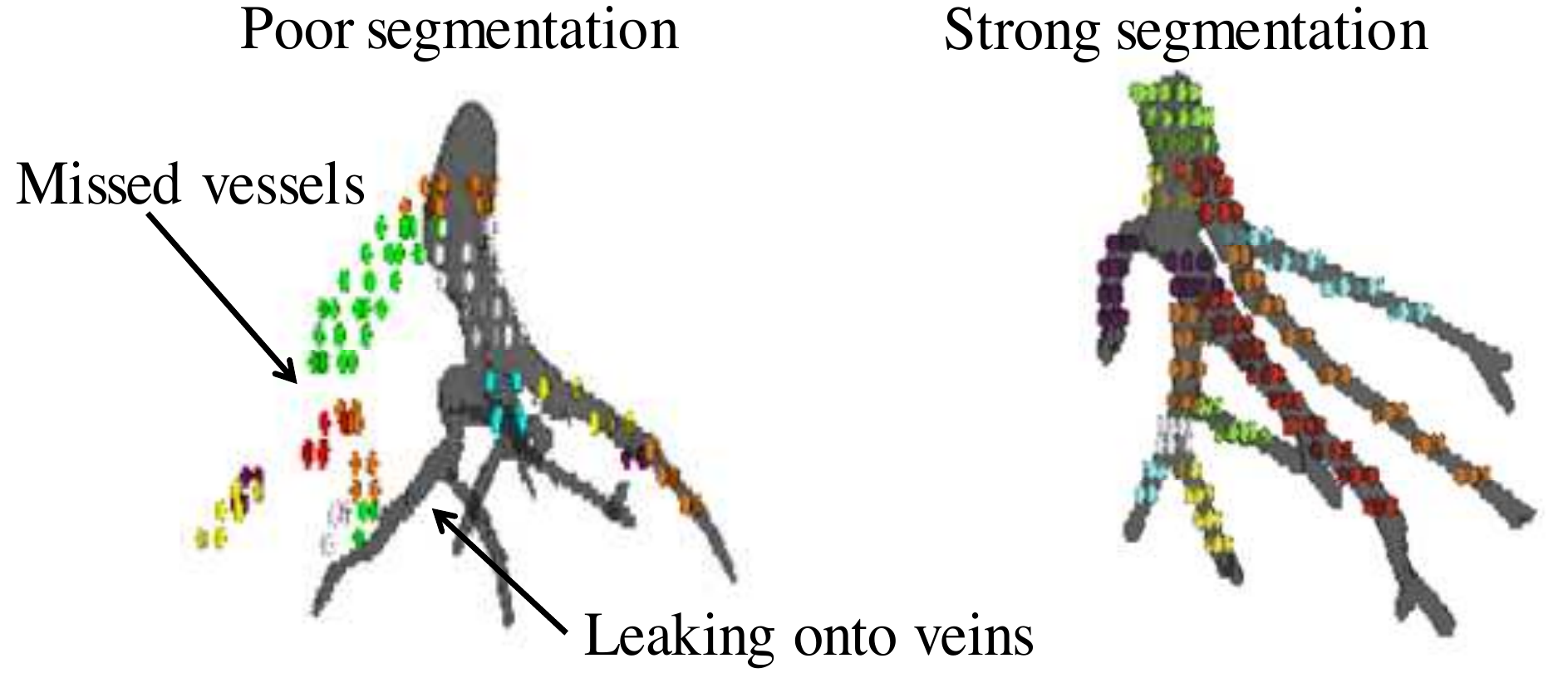} 
\end{array}$
\end{center}
\caption{3D representation of the artery segmentation results, overlaid on the manually placed SS ground truth points, shown for a strong segmentation and an unusually weak segmentation. In the weak case, failure of the bifurcation detector early in the tracking resulted in missing several true arteries (false-negatives).}
\label{fig:cases}
\end{figure}

Figure~\ref{fig:cases} shows the results of the weakest and strongest segmentations, overlaid with truth markings for the arteries. In the case of the strong segmentation, the algorithm was able to correctly seed the pulmonary artery, and separate and segment the arterial tree to four generations. Additionally, the true vessel surface matched the segmented surface very closely. In the case of the weak segmentation, a segmental child branch failed to be detected, which resulted in several missed arteries. The vessel tracker also leaked onto the adjacent venous tree, returning false-positives (non-arteries). 

\begin{figure}[!t] \centering \includegraphics[width=5.2in]{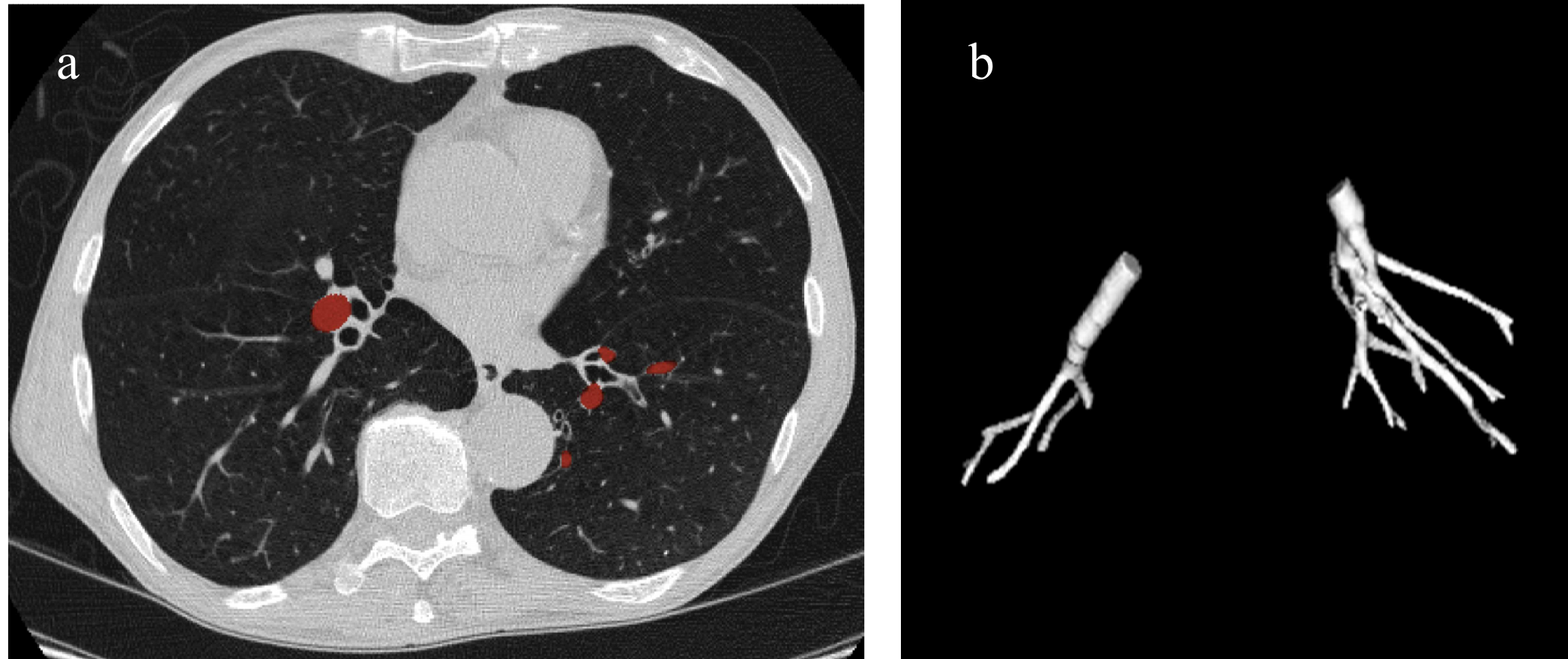} \caption{a) Axial slice from a section of the inferior lung. Vessels in red indicate the segmentation result. b) Three dimensional rendering of the pulmonary arterial tree generated by our automated method. } \label{fig:results} \end{figure}



\begin{figure*}[!t] \centering \includegraphics[width=5.5in]{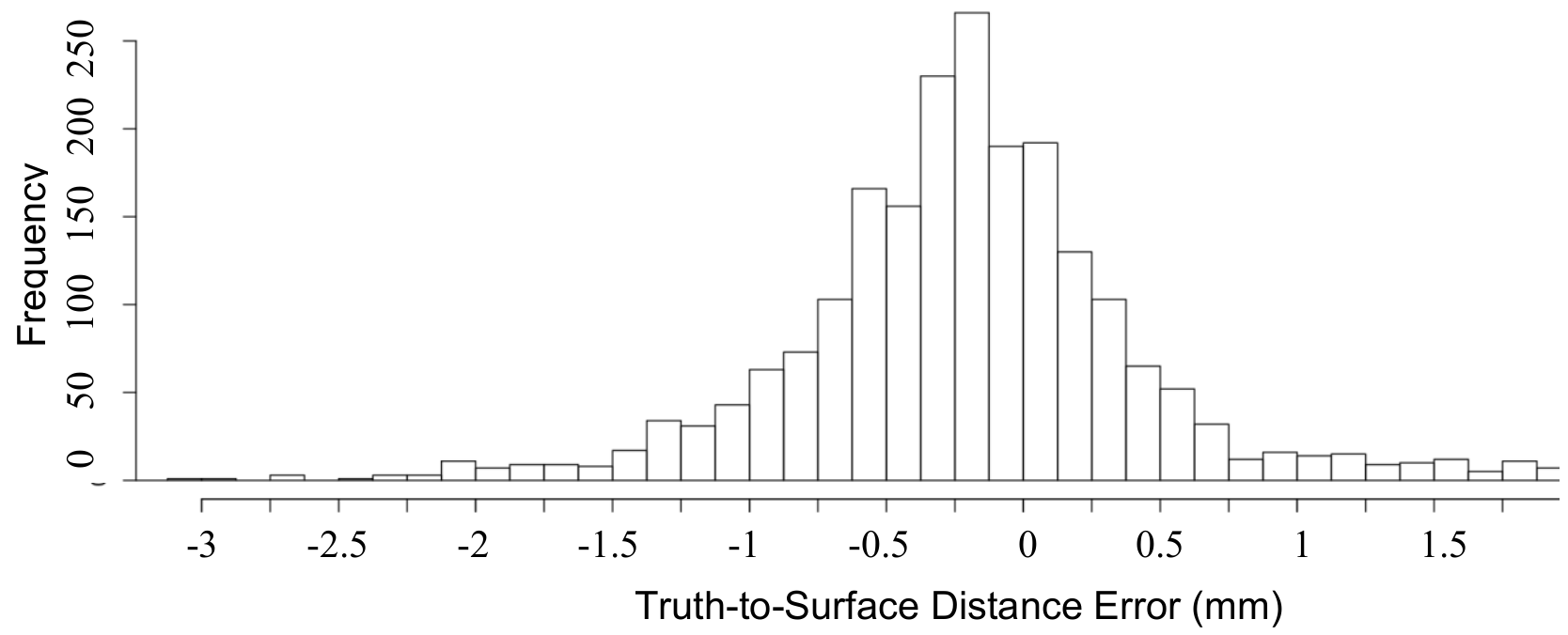} \caption{Distribution of SS metric distance errors for correctly identified vessels using sparse markings across 10 cases. The values represent the Euclidean distance between the manual truth points and the nearest surface point. The averaged signed distance error was -0.15 mm (~1/5 voxel).} \label{fig:dist} \end{figure*}

Using the SS metric, we computed an overall under-segmentation bias of 0.15 mm, roughly 1/5 of a voxel. The RMS distance error between the manual markings and the segmentation surface was 0.63 mm, less than one voxel. Among identified vessels, the greatest under-segmentation error was 5.25 mm, which corresponded to a vessel segmentation that terminated before the entire true vessel was segmented. The greatest over-segmentation was 3.14 mm. This point was found near the lung hilum, where the arteries and veins are closely packed, resulting in minor over-segmentation. Figure~\ref{fig:dist} shows the overall distribution of SS metric distance errors for the segmented vessels. 

\subsection{Sparse surface metric}

\begin{figure}[h]
\begin{center}$
\begin{array}{cc}
\includegraphics[width=5.3in]{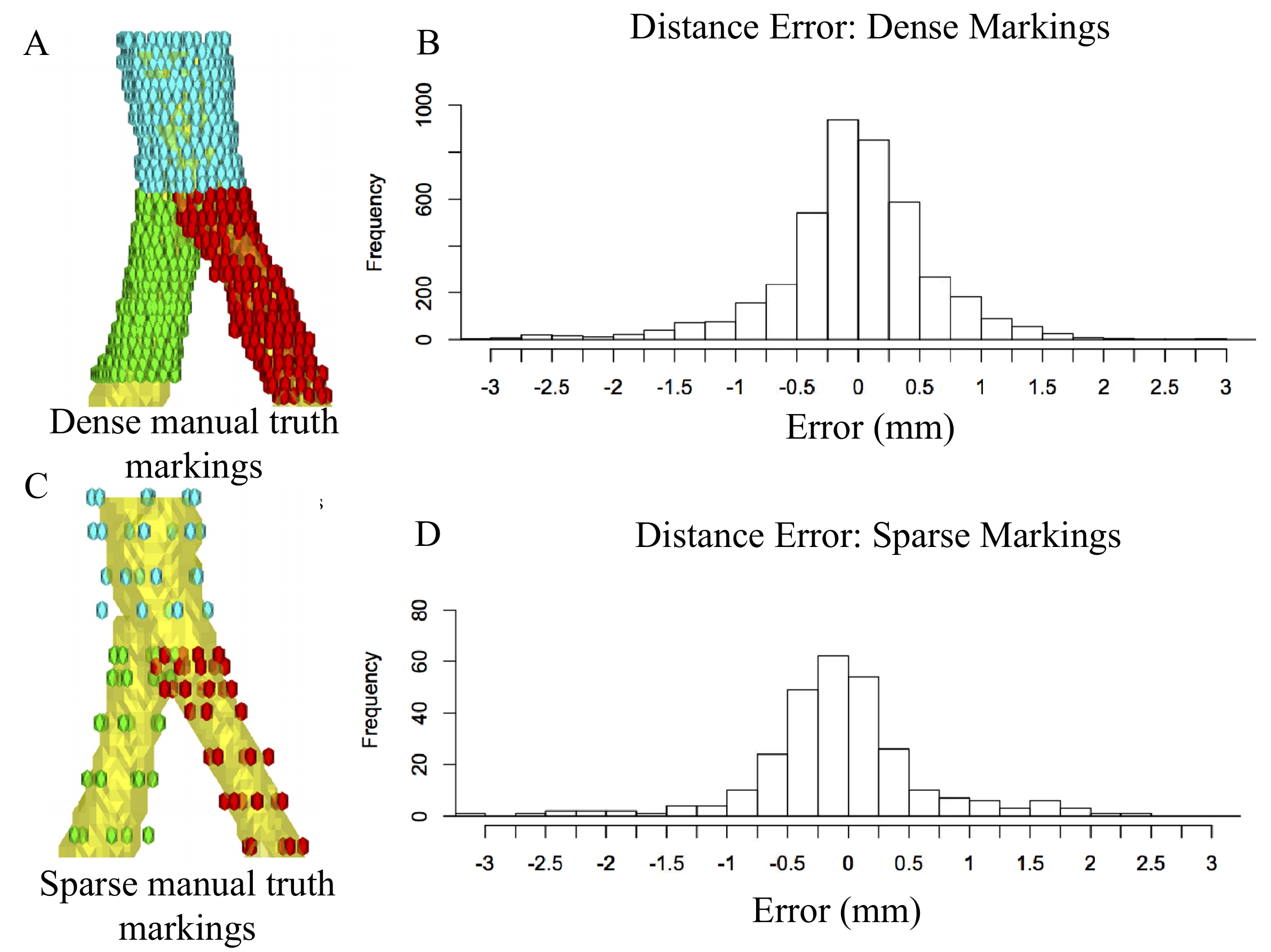} 
\end{array}$
\end{center}
\caption{Manual SS markings overlaid on the left main pulmonary artery and its two child branches}
\label{fig:ssm}
\end{figure}

To validate the use of sparse surface markings, we marked a single case with both dense and sparse coverage, and applied the SS metric to each truth set for validating the segmentation. The dense manual markings, shown in Figure~\ref{fig:ssm}a overlaid on a small section of the segmentation result, represent the most ideal coverage of the true vessel surface. In total, 4780 points were manually marked to completely cover the pulmonary arteries and veins of a single case. Figure~\ref{fig:ssm}b shows the same vessel covered with sparse truth markings. For sparse coverage, 302 points were applied to cover the same arteries and veins marked under dense coverage. We then asked the question: Under the SS metric, does sparse coverage yield the same validation information as dense coverage?

The results of the SS metric distance errors are shown in Figure~\ref{fig:ssm}. The ASD error calculated using the dense coverage was 0.29 mm, while the ASD error using the sparse coverage was 0.27 mm. Using an unpaired t-test, the 95\% confidence interval for the difference in means between the two samples was (-0.07, 0.10). Thus, we find that under the SS metric, sparse marking of the ground truth is equivalent to more labor-intensive dense markings. 

\section{Discussion}

Here we have presented a method for 1) automatically locating seed points within pulmonary arteries, 2) tracking the seeded arteries into the lungs while maintaing the separation between the arterial and venous trees, and 3) quantitatively validating the results using a novel evaluation metric. The vessel tracking begins near the hilum of the left and right lungs, and uses a model-based bifurcation detector to segment the child vessels. In using a model-based method, we were able to maintain separation between the isointensity arterial and venous trees (90\% specificity for arteries) based on predictable branching geometries. The segmentation was performed in binary space, which reduced both the signal from the vessels and the noise inherent to low dose scans. The arteries were separated from the veins through vessel morphology and relative anatomical location alone. This approach is best suited for scans without contrast enhancement, where vessel edge information is lacking. 

The vessel tracking algorithm was able to successfully segment vessels of greatly different radii. By fitting the size of the model cylinders to the vessel tree, the tracker was able to segment both the large parent arteries of the inferior lobes and the small subsegmental arteries deep in the lung parenchyma. However, as the vessel radii approached the smallest possible cylinder, the tracker was unable to detect bifurcations. The method was strongest in the larger vessels where Murray's law of vessel bifurcation is more closely followed. As vessel sizes decrease, blood flow becomes more non-Newtonian, and child vessel sizes become larger relative to the parent. Additionally, the smaller vessels deep in the lung parenchyma were more prone to obfuscation by noise because of their small size. 

In most cases, the tracker was able to separate the arteries from the veins by searching for cylindrical objects in a limited size and orientation search space defined by the bifurcation model. While this approach was able to distinguish nearly identical vessels based only on their relationship to other arteries, it is unable to locate vessels that fall outside of the search space. Thus, arteries were missed if they originated at apparent trifurcations, or had unusually small radii relative to their sister branches. Relaxing the bifurcation model results in few missed segments, but more false positives (i.e. segmented veins). 

The main source of error in vessel segmentation occurred at the left and right inferior pulmonary veins. In cases where the arteries were highly conflated with the veins and airway walls, due either to high noise or unusual anatomy, the vessel tracker leaked onto the pulmonary veins and tracked them into the parenchyma. One possible method for reducing the false positive rate would be to initiate a vessel tracker at the main inferior pulmonary arteries, and exclude from the arterial segmentation vessels that were highly likely to belong to the venous tree. 

Automated detection of parent vessels in the left and right upper lobes is the next appropriate step to take in the future. In this work, only the largest arteries to enter the lung parenchyma were identified. Such a method could employ a similar ROI construction or use another algorithm to locate the upper lobe arteries. The sensitivity of the algorithm could be improved by developing a trifurcation detector, or developing a method for identifying small subsegmental arteries missed by the tracker. Additionally, the method could be paired with a global Hessian matrix filter to amplify the signal from vessels and reduce tracking error.

Validation of complex structures is a challenging task, and a novel surface-based ground truth marking system and validation metric were defined for use with complex structures. Sparse surface markings capture significant vessel detail while require fewer markings that traditional dense surface markings. This allows for validation on a greater number of cases given the same amount of effort. The sparse surface markings were shown to be an accurate sampling of the true surface, justifying the use of spares markings over dense markings. 




\section{Conclusion}

Pulmonary artery segmentation has been suggested as a method to improve diagnosis of both pulmonary embolism and pulmonary arterial hypertension. Although much work has focused on segmenting the arteries and veins as one structure, very little previous work has been performed to distinguish the two trees. In this work, a method was developed for automatically segmenting the pulmonary arteries in low-dose CT, and was applied to the basal pulmonary arteries and their child vessels.

\bibliographystyle{model1-num-names}
\bibliography{Wala2011BIB}

\begin{thebibliography}{15}
\expandafter\ifx\csname natexlab\endcsname\relax\def\natexlab#1{#1}\fi
\providecommand{\bibinfo}[2]{#2}
\ifx\xfnm\relax \def\xfnm[#1]{\unskip,\space#1}\fi
\bibitem[{Croisille et~al.(1995)Croisille, Souto, Cova, Wood, Afework, Kuhlman,
  and Zerhouni}]{Croisille:1995}
\bibinfo{author}{P.~Croisille}, \bibinfo{author}{M.~Souto},
  \bibinfo{author}{M.~Cova}, \bibinfo{author}{S.~Wood},
  \bibinfo{author}{Y.~Afework}, \bibinfo{author}{J.~Kuhlman},
  \bibinfo{author}{E.~Zerhouni},
\newblock \bibinfo{title}{{Pulmonary nodules: improved detection with vascular
  segmentation and extraction with spiral CT}},
\newblock \bibinfo{journal}{Radiology} \bibinfo{volume}{197}
  (\bibinfo{year}{1995}) \bibinfo{pages}{397--401}.
\bibitem[{Masutani et~al.(2002)Masutani, MacMahon, and Doi}]{Masutani:2002}
\bibinfo{author}{Y.~Masutani}, \bibinfo{author}{H.~MacMahon},
  \bibinfo{author}{K.~Doi},
\newblock \bibinfo{title}{{Computerized detection of pulmonary embolism in
  spiral CT angiography based on volumetric image analysis}},
\newblock \bibinfo{journal}{IEEE Transaction on Medical Imaging}
  \bibinfo{volume}{21} (\bibinfo{year}{2002}) \bibinfo{pages}{1517--1523}.
\bibitem[{Linguraru et~al.(2010)Linguraru, Pura, Uitert, Mukherjee, Summers,
  Minniti, Gladwin, Kato, Machado, and Wood}]{Linguraru:2010}
\bibinfo{author}{M.~Linguraru}, \bibinfo{author}{J.~Pura},
  \bibinfo{author}{R.~V. Uitert}, \bibinfo{author}{N.~Mukherjee},
  \bibinfo{author}{R.~Summers}, \bibinfo{author}{C.~Minniti},
  \bibinfo{author}{M.~Gladwin}, \bibinfo{author}{G.~Kato},
  \bibinfo{author}{R.~Machado}, \bibinfo{author}{R.~Wood},
\newblock \bibinfo{title}{{Segmentation and quantification of pulmonary artery
  for noninvasive CT assessment of sickle cell secondary pulmonary
  hypertension}},
\newblock \bibinfo{journal}{Medical Physics} \bibinfo{volume}{37}
  (\bibinfo{year}{2010}) \bibinfo{pages}{1522--1532}.
\bibitem[{Lee and Reeves(2009)}]{LeeJ:2008}
\bibinfo{author}{J.~Lee}, \bibinfo{author}{A.~Reeves},
\newblock \bibinfo{title}{{Segmentation of the airway tree from chest CT using
  local volume of interest}},
\newblock \bibinfo{journal}{Second International Workshop of Pulmonary Image
  Analysis}  (\bibinfo{year}{2009}) \bibinfo{pages}{353--364}.
\bibitem[{B\"ulow et~al.(2005)B\"ulow, Wiemker, Blaffert, Lorenz, and
  Renisch}]{Bulow:2005}
\bibinfo{author}{T.~B\"ulow}, \bibinfo{author}{R.~Wiemker},
  \bibinfo{author}{T.~Blaffert}, \bibinfo{author}{C.~Lorenz},
  \bibinfo{author}{S.~Renisch},
\newblock \bibinfo{title}{{Automatic extraction of the pulmonary artery tree
  from multi-slice CT data}},
\newblock \bibinfo{journal}{Medical Imaging 2005: Proceedings of SPIE}
  \bibinfo{volume}{5746} (\bibinfo{year}{2005}) \bibinfo{pages}{730--740}.
\bibitem[{Susan~Wood et~al.(1995)Susan~Wood, Hoford, Hoffman, and
  Mitzner}]{Wood:1995}
\bibinfo{author}{E.~Z. Susan~Wood}, \bibinfo{author}{J.~Hoford},
  \bibinfo{author}{E.~Hoffman}, \bibinfo{author}{W.~Mitzner},
\newblock \bibinfo{title}{{Measurement of three-dimension lung tree structures
  by using computed tomography}},
\newblock \bibinfo{journal}{Journal of Applied Physiology} \bibinfo{volume}{79}
  (\bibinfo{year}{1995}) \bibinfo{pages}{1687--1697}.
\bibitem[{Masutani et~al.(2001)Masutani, MacMahon, and Doi}]{Masutani:2001}
\bibinfo{author}{Y.~Masutani}, \bibinfo{author}{H.~MacMahon},
  \bibinfo{author}{K.~Doi},
\newblock \bibinfo{title}{{Automated segementation and visualization of the
  pulmonary vascular tree in spiral CT angiography: an anatomy-oriented
  approach based on three-dimensional image analysis}},
\newblock \bibinfo{journal}{Journal of Computer Assisted Tomography}
  \bibinfo{volume}{25} (\bibinfo{year}{2001}) \bibinfo{pages}{587--597}.
\bibitem[{Zhou et~al.(2007)Zhou, Chan, Sahiner, Hadjiski, and
  Chughtai}]{Zhou:2007}
\bibinfo{author}{C.~Zhou}, \bibinfo{author}{H.-P. Chan},
  \bibinfo{author}{B.~Sahiner}, \bibinfo{author}{L.~Hadjiski},
  \bibinfo{author}{A.~Chughtai},
\newblock \bibinfo{title}{{Automatic multiscale enhancement and segmentation of
  pulmonary vessels in CT pulmonary angiography images for CAD applications}},
\newblock \bibinfo{journal}{Medical Physics} \bibinfo{volume}{34}
  (\bibinfo{year}{2007}) \bibinfo{pages}{4567--4577}.
\bibitem[{Shikata et~al.(2009)Shikata, McLennan, Hoffman, and
  Sonka}]{Shikata:2009}
\bibinfo{author}{H.~Shikata}, \bibinfo{author}{G.~McLennan},
  \bibinfo{author}{E.~Hoffman}, \bibinfo{author}{M.~Sonka},
\newblock \bibinfo{title}{{Segmentation of pulmonary vascular trees from
  thoracic 3D CT images}},
\newblock \bibinfo{journal}{International Journal of Biomedical Imaging}
  \bibinfo{volume}{2009} (\bibinfo{year}{2009}).
\bibitem[{Zhou et~al.(2007)Zhou, Chang, Metaxas, and Axel}]{ZhouJ:2007}
\bibinfo{author}{J.~Zhou}, \bibinfo{author}{S.~Chang},
  \bibinfo{author}{D.~Metaxas}, \bibinfo{author}{L.~Axel},
\newblock \bibinfo{title}{{Vascular structure segmentation and bifurcation
  detection}},
\newblock \bibinfo{journal}{IEEE International Symposium on Biomedical Imaging}
   (\bibinfo{year}{2007}) \bibinfo{pages}{872--875}.
\bibitem[{Kaftan et~al.(2008)Kaftan, Bakai, Das, and Aach}]{Kaftan_2:2008}
\bibinfo{author}{J.~N. Kaftan}, \bibinfo{author}{A.~Bakai},
  \bibinfo{author}{M.~Das}, \bibinfo{author}{T.~Aach},
\newblock \bibinfo{title}{{Locally adaptive fuzzy pulmonary vessel segmentation
  in contrast enhanced CT}},
\newblock \bibinfo{journal}{Proceedings 5th international symposium on
  biomedical imaging: from nano to macro}  (\bibinfo{year}{2008})
  \bibinfo{pages}{101--104}.
\bibitem[{Lei et~al.(2001)Lei, Udupa, Saha, and Odhner}]{Lei:2001}
\bibinfo{author}{T.~Lei}, \bibinfo{author}{J.~Udupa},
  \bibinfo{author}{P.~Saha}, \bibinfo{author}{D.~Odhner},
\newblock \bibinfo{title}{{Artery-vein separation via MRA - An image processing
  approach}},
\newblock \bibinfo{journal}{IEEE Transaction on Medical Imaging}
  \bibinfo{volume}{20} (\bibinfo{year}{2001}) \bibinfo{pages}{689--703}.
\bibitem[{van Bemmel et~al.(2003)van Bemmel, Spreeuwers, Viergever, and
  Niessen}]{Bemmel:2003}
\bibinfo{author}{C.~van Bemmel}, \bibinfo{author}{L.~Spreeuwers},
  \bibinfo{author}{M.~Viergever}, \bibinfo{author}{W.~Niessen},
\newblock \bibinfo{title}{{Level-set-based artery-vein separation in blood pool
  agent CE-MR angiograms}},
\newblock \bibinfo{journal}{IEEE Transactions on Medical Imaging}
  \bibinfo{volume}{22} (\bibinfo{year}{2003}) \bibinfo{pages}{1224--1234}.
\bibitem[{Saha et~al.(2010)Saha, Gao, Alford, and Sonka}]{Saha:2010}
\bibinfo{author}{P.~Saha}, \bibinfo{author}{Z.~Gao},
  \bibinfo{author}{S.~Alford}, \bibinfo{author}{M.~Sonka},
\newblock \bibinfo{title}{{Topomorphological separation of fused isointensity
  objects via multiscale opening: separating arteries and veins in 3-D
  pulmonary CT}},
\newblock \bibinfo{journal}{IEEE Transaction on Medical Imaging}
  \bibinfo{volume}{29} (\bibinfo{year}{2010}) \bibinfo{pages}{840--851}.
\bibitem[{Lee and Reeves(2009)}]{Lee:2009}
\bibinfo{author}{J.~Lee}, \bibinfo{author}{A.~Reeves},
\newblock \bibinfo{title}{{Segmentation of the airway tree from chest CT using
  local volume of interest}},
\newblock \bibinfo{journal}{Second International Workshop of Pulmonary Image
  Analysis}  (\bibinfo{year}{2009}) \bibinfo{pages}{353--364}.

\end{thebibliography}







\end{document}